\def\year{2017}\relax
\documentclass[letterpaper]{article}
\usepackage{aaai17}
\usepackage{times}
\usepackage{helvet}
\usepackage{courier}
\newcommand{\etal}{\mbox{\emph{et al.\ }}}
\newcommand{\eg}{\mbox{\emph{e. g.\ }}}
\newcommand{\ie}{\mbox{\emph{i. e.\ }}}

\usepackage{multirow}
\usepackage{graphicx}
\usepackage{amsmath,amssymb} 
\usepackage{mathrsfs}

\begin{document}
%
\title{A Multi-task Deep Network for Person Re-identification}
\author{Weihua Chen$^{1}$, Xiaotang Chen$^{1}$, Jianguo Zhang$^{3}$, Kaiqi Huang$^{1,2,4}$\\
$^{1}$CRIPAC$ \& $NLPR, CASIA \quad $^{2}$University of Chinese Academy of Sciences\\
$^{3}$Computing, School of Science and Engineering, University of Dundee, United Kingdom\\
$^{4}$CAS Center for Excellence in Brain Science and Intelligence Technology\\
\normalsize{Email:$\{$weihua.chen, xtchen, kqhuang$\}$@nlpr.ia.ac.cn, j.n.zhang@dundee.ac.uk}}

\maketitle
\begin{abstract}
Person re-identification (ReID) focuses on identifying people across different scenes in video surveillance, which is usually formulated as a binary classification task or a ranking task in current person ReID approaches. In this paper, we take both tasks into account and propose a multi-task deep network (MTDnet) that makes use of their own advantages and jointly optimize the two tasks simultaneously for person ReID. To the best of our knowledge, we are the first to integrate both tasks in one network to solve the person ReID. We show that our proposed architecture significantly boosts the performance. Furthermore, deep architecture in general requires a sufficient dataset for training, which is usually not met in person ReID. To cope with this situation, we further extend the MTDnet and propose a cross-domain architecture that is capable of using an auxiliary set to assist training on small target sets. In the experiments, our approach outperforms most of existing person ReID algorithms on representative datasets including CUHK03, CUHK01, VIPeR, iLIDS and PRID2011, which clearly demonstrates the effectiveness of the proposed approach.
\end{abstract}

\section{Introduction}


Person re-identification (ReID) is an important task in wide area video surveillance.
The key challenge is the large appearance variations,
usually caused by the significant changes in human body poses, illumination and camera views.
It has many applications, such as inter-camera pedestrian tracking and human retrieval.

\begin{figure*}
\centering
\includegraphics[width=1\linewidth]{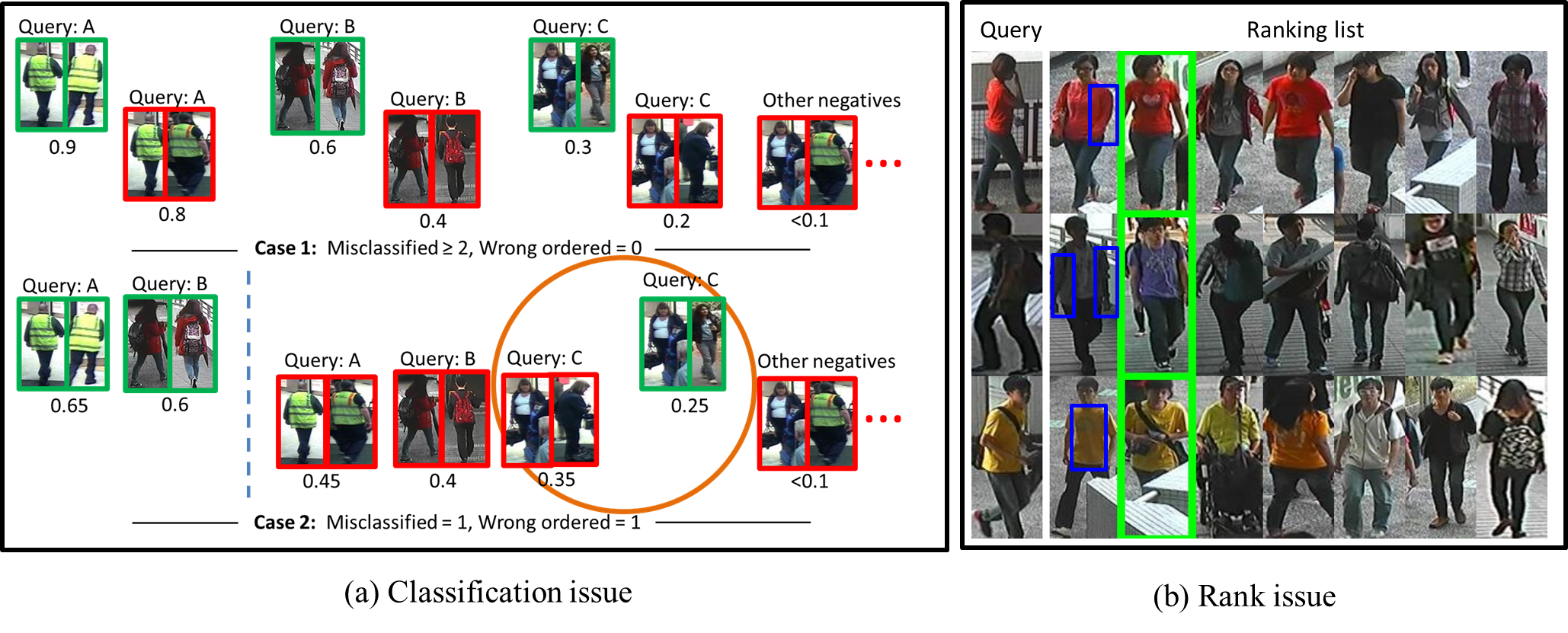}
\caption{Problems in two tasks.(a) Classification issue: the classification loss prefer to train a lower misclassification rate model like Case 2 rather than Case 1.
(b) Ranking issue: the appearance of top-rank images is more similar to the query image, while the true positive presents a much less similar appearance. (Best viewed in color and see main text for detailed explanation)}
\label{fig:twoissue}
\end{figure*}


Recently, deep learning approaches~\cite{Li/cvpr2014deepreid,ahmed/cvpr2015improved,sircir2016cvpr} are successfully employed in person ReID with significant performance, especially on large datasets, such as CUHK03.
Most deep learning methods~\cite{Li/cvpr2014deepreid,yi/icpr2014deep,ahmed/cvpr2015improved} solve the problem as a binary classification issue and adopt a classification loss (\eg a softmax loss) to train their models.
The core behind these approaches is to learn identifiable features for each pair for classification.
The binary classification loss is usually designed to require all positive pairs should hold smaller distances than all negative pairs.
However, in person ReID, we don't have to require all positive pairs holding smaller distances
than all negative pairs \textit{regardless of query images}.
Instead, what we want is \textit{for each query image}, its positive pairs have smaller distances than its negative ones.
Therefore, in some cases\footnote{This situation commonly happens when a fixed embedding metric, \eg Euclidean distance, is used for similarity measurement. In this case, it's hard for the network to learn a suitable feature representation.},
the application of binary classification loss may lead the learned model to an undesired locally optimal solution, which is elaborated as below.

The example is shown in Fig.~\ref{fig:twoissue} (a). Case 1 and 2 illustrate two projected distributions of scores obtained by trained binary classifiers. For each pair sample, the score underneath denotes the similarity probability between its two images. Query:\textit{X} indicates where an image from person \textit{X} is used as a query image (the left image in a pair). For example, Query:A means an image from person A is used as a query image. Green-coloured rectangle indicates a positive pair, and red rectangle for the negative pair. In Case 1, it is evident that for each query image (w.r.t one particular person), we can get the correct rank-1 match, \ie two images within its positive pairs always hold larger similarity score than those within its negative pairs. However, in this case it is very difficult for a classifier to determine a suitable threshold to get a low misclassification cost (\eg less than two misclassified samples). On the contrary in Case 2, where the vertical dashed line denotes the decision threshold learned by the classifier, the classifier has a lower misclassification rate. As a result, a binary classifier will favor Case 2 rather than Case 1, as the classification loss in Case 2 will be lower than that in Case 1. But in ReID, we prefer Case 1, which outputs correct ranking results for all of the three persons, rather than Case 2 that contains a false rank-1 result (highlighted in an orange circle). Case 2 could be potentially rectified by a ranking loss.


As person ReID commonly uses the Cumulative Matching Characteristic (CMC) curve for performance evaluation which follows rank-\emph{n} criteria, some deep learning approaches~\cite{ding2015deep,chen2015deep,imptrp2016cvpr} begin to treat the person ReID as a ranking task, similar to image retrieval, and apply a ranking loss (\eg a triplet loss) to address the problem. The main purpose is to keep the positive pairs maintaining shorter relative distances in the projected space.
However, the person ReID differs from image retrieval in that person ReID needs to identify the same person across different scenes (\ie, a task of predicting positive and negative pairs, focusing on identifiable feature learning, and a positive pair is not necessarily the most similar pair in appearance).
Ranking-based approaches are sensitive to their similarity measurements.
The current measurements (\eg the Euclidean distance in the triplet loss) care more about the similarity to query images in appearance.
In the projection space obtained by a model trained on the triplet loss, it's very challenging to find out a true positive which holds a less similar appearance.
As shown in Fig.~\ref{fig:twoissue} (b), there are three query images.
Each has a ranking list returned by a ranking loss, and the left-most is the most similar one to the query.
The green rectangle indicates the positive pair (\textit{ground truth}).
We can observe that the image ranked first w.r.t each query image is a mismatched image but holding a more similar appearance to the query image than the matched does.

In the person ReID, either the binary classification loss or the ranking loss has its own strengths and weaknesses.
As two tasks handle the person ReID from different aspects,
we take both of them into account and build a more comprehensive person ReID algorithm.
In our method, two tasks are jointly optimized in one deep network simultaneously.
We set the binary classification loss and the ranking loss on different layers according their own advantages.
The ranking loss encourages a relative distance constraint, while the classification loss seeks to learn discriminative features for each pair during the similarity measurement.
As the classification task focuses on feature of pairs, we import the joint feature maps to represent the relationships of paired person images.

Meanwhile, deep learning approaches, such as convolutional neural networks (CNN), benefit a lot from a large scale dataset (\eg ImageNet). However, this is not the case in person ReID.
Since manually labeling image pairs is tedious and time-consuming, most of current ReID datasets are often of limited sizes,
\eg CUHK01~\cite{li/accv2012human}, VIPeR~\cite{gray2007evaluating}, iLIDS~\cite{zheng/bmvc2009associating} and PRID2011~\cite{hirzer2011person}. It could hinder the attempts to maximize the learning potential of our proposed network on each of those datasets. This case can be migrated by using some auxiliary datasets.
However, the variations across camera views are different from dataset to dataset.
As a consequence, the data of the auxiliary dataset can't be directly used to train models on small datasets.
In this paper, the problem is considered as a semi-supervised cross-domain issue~\cite{ganin2014unsupervised}.
The target domain is the small dataset that contains only a few samples
and the source domain is an auxiliary dataset which is large enough for training CNN models.
As person ReID can be considered as a binary classification problem,
our purpose is to keep the samples of the same class in different domains closer.
A cross-domain architecture is further proposed to minimize the difference of the joint feature maps in two datasets,
which are belonged to the same class of pairs (\ie, positive pair and negative pair),
and utilize the joint feature maps of the auxiliary dataset to fine tune those of small datasets during the training process.
In this case, the joint feature maps of small datasets are improved with the data of the auxiliary dataset
and boost the ReID performance on smaller target datasets.

In summary, our contributions are three-fold:
1) a novel multi-task deep network for person ReID, where two tasks focuses on different layers and are jointly optimized simultaneously for person ReID;
2) a cross-domain architecture based on the joint feature maps to handle the challenge of limited training set;
3) a comprehensive evaluation of our methods on five datasets, and showing the superior performance over most of state-of-the-art methods.

\section{Related work}
\label{sec:relatedwork}

Most of existing methods in person ReID focus on either feature extraction
\cite{Zhao/cvpr2014midlevel,su2015multi,hierarchical2016cvpr}, or similarity measurement
\cite{Li/cvpr2013locally,Shen/iccv2015structure,Liao/iccv2015psd}.
Person image descriptors commonly used include color histogram \cite{Koestinger/cvpr2012scale,Li/cvpr2013locally,xiong/eccv2014person},
local binary patterns \cite{Koestinger/cvpr2012scale},
Gabor features \cite{Li/cvpr2013locally}, and etc.,
which show certain robustness to the variations of poses, illumination and viewpoints.
For similarity measurement, many metric learning approaches are proposed to learn a suitable metric,
such as locally adaptive decision functions \cite{li/cvpr2013learning},
local fisher discriminant analysis \cite{pedagadi/cvpr2013local},
cross-view quadratic discriminant analysis \cite{liao2015person},
and etc.
A few of them \cite{xiong/eccv2014person,paisitkriangkrai/cvpr2015learning} learn a combination of multiple metrics.
However, manually crafting features and metrics require empirical knowledge,
and are usually not optimal to cope with large intra-person variations.

Since feature extraction and similarity measurement are independent,
the performance of the whole system is often suboptimal
compared with an end-to-end system using CNN that can be globally optimized via back-propagation.
With the development of deep learning and increasing availability of datasets,
the handcrafted features and metrics struggle to keep top performance widely, especially on large scale datasets.
Alternatively, deep learning
is attempted for person ReID to automatically learn features and metrics
\cite{Li/cvpr2014deepreid,ahmed/cvpr2015improved,sircir2016cvpr}.
Some of them~\cite{ding2015deep,chen2015deep,imptrp2016cvpr} consider person ReID as a ranking issue.
For example, Ding \etal~\cite{ding2015deep} use a triplet loss to get the relative distance between images.
Chen \etal~\cite{chen2015deep} design a ranking loss which minimizes the cost corresponding to the sum
of the gallery ranking disorders.
Cheng \etal~\cite{imptrp2016cvpr} add a new term to the original triplet loss function to further constrain the distances
of pairs.

Other approaches~\cite{Li/cvpr2014deepreid,ahmed/cvpr2015improved,wu2016enhanced} tackle the person ReID problem from the classification aspect.
For instance, Yi \etal~\cite{yi/icpr2014deep} utilize a siamese convolutional neural network to train a feature representation.
Li \etal \cite{Li/cvpr2014deepreid} design a deep filter pairing neural network to solve the ReID problem.
Ahmed \etal \cite{ahmed/cvpr2015improved} employ a local neighborhood difference to deal with this misalignment issue.
All of them employ a binary classification loss to train their models.
It is worth mentioning that there are some papers~\cite{wu2016enhanced,dgd2016cvpr} using multi-class classification instead of binary classification. They classify identities to solve the person ReID problem, which shares a similar idea with DeepID in face recognition~\cite{sun2014deep}. However, in most person ReID datasets, there are few samples for each identity.
VIPeR~\cite{gray2007evaluating} and PRID2011~\cite{hirzer2011person} datasets have only two images for each person.
The lack of training samples may make the multi-class classification less effective.
Xiao \etal~\cite{dgd2016cvpr} achieve a good performance, but it combines all current datasets together as its training data.

Our network considers two tasks (the classification loss and the ranking loss) simultaneously and takes both of their advantages during training.
Wang \etal~\cite{sircir2016cvpr} also discuss both classification and ranking losses,
however, it trains two losses separately and combines them on the score level.
In this paper, we jointly optimize two tasks simultaneously in our network.

It is worth noting that none of the works above in person ReID seeks to solve the problem of ``learning a deep net on a small dataset'' which is a typical case in person ReID. This paper addresses this issue by proposing a cross-domain deep architecture capable of learning across ReID datasets.

\begin{figure}
\centering
\includegraphics[width=1.0\linewidth]{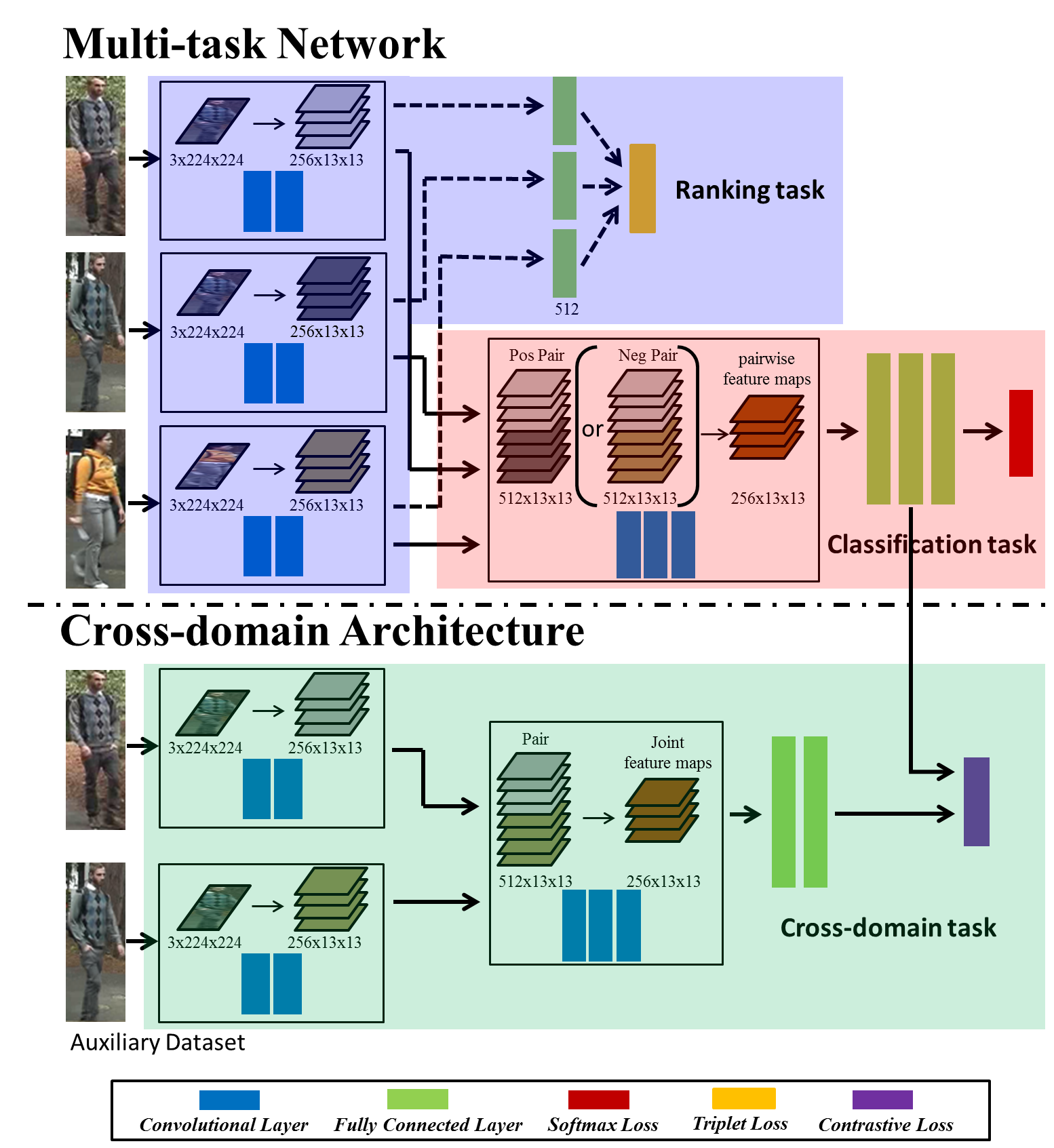}
\caption{The framework of the proposed multi-task deep network and the cross-domain architecture.
The cross-domain architecture is only used when an auxiliary dataset is needed for training.}
\label{fig:framework}
\end{figure}

\section{The proposed network}
\label{sec:cnnarchitecture}

\subsection{The multi-task network}
\label{ssec:multitask}

In our method, we build our architecture according to the different focuses of two tasks.
As we known, the ranking task concentrates on the orders of images with the same query.
Its purpose is to rank the similarities of images and obtain a good ranking list for each query.
For two person images, in order to compute their similarity score, we have to compare each part of two people.
We can't obtain their similarity score only based on some local parts.
In other words, the global features of the whole images should be paid more attention than local parts during ranking~\cite{tolias2015particular}.
Meanwhile, in the association, the most important purpose of the classification task is to distinguish two categories
and make the learned features more identifiable.
As shown in Fig.~\ref{fig:twoissue} (b),
the possible key to distinguish the top 1 result from the query is mainly on the blue local regions,
\eg using the feature of the sleeves or the belting.
So the classification loss should pay more attention on learning these local semantic features,
which hold enough identifiable information.
In this way, the classification loss would prefer to semantic local features instead of the global features during training.

From Wang's work~\cite{wang2015visual}, it had been shown that the higher layers in deep network capture semantic concepts,
whereas lower layers encode features to capture intra-class variations.
For ranking, we compare images based on a combination (global appearance oriented) of low-level features (\ie edges, bars etc) learned in lower layers to overcome intra-class variations (as suggested by Wang's work~\cite{wang2015visual}). Features in high layers focus on identifiable local semantic concepts, driven by the classification loss.
The whole framework is shown in Fig.~\ref{fig:framework}.
The ranking loss provides global low-level features which could be appropriate for image similarity ranking,
and the classification loss further learns the identifiable local features based on the low-level ones.
Then we give the details of our multi-task network.

The ranking part is a triplet-input model. For each positive pair, we produce ten triplets (a positive pair + a negative image: $A_1,A_2,B_2$ \footnote{$A,B$ are the person IDs and $1,2$ mean the camera IDs.}).
All these triplets constitute our training data.
The input triplet contains three images, each of the size $3*224*224$.
The ranking task includes two convolutional layers at the beginning, which are used to reinforce the learning of global features.
After the two convolutional layers, three sets of feature maps hold the same size of $256*13*13$
and are sent to a triplet loss through a shared fully connected layer.
The triplet loss being minimized is the same as FaceNet~\cite{schroff2015facenet}:

\begin{equation}
L_{trp}=\sum_{i=1}^N{[\|f_{A1}-f_{A2}\|_2^2-\|f_{A1}-f_{B2}\|_2^2+\alpha]_{+}}
\label{eq:triploss}
\end{equation}

where $\alpha$ is a margin that is enforced between positive and negative pairs,
$N$ is the number of the triplets. $f\in \mathbb{R}^{512}$ denotes the features input to the triplet loss from three images.
Minimizing the triplet loss is to reserve the information of relative distances between input images.

In the classification part, the input of the third convolutional layer is a set of feature maps of an image pair.
The three sets of feature maps with the size of $256*13*13$ from the ranking task are regrouped into two types of pairs,
a positive pair and a negative pair.
The feature maps from the two images of the same person, \ie ($A_1,A_2$), are concatenated as a positive pair,
while one image in the positive pair ($A_1$) and one negative image ($B_2$) from the different camera view are stacked to form the negative pair.
The size of feature maps of each pair is $512*13*13$.
These two pairs are fed to three convolutional layers in order, one at each time.
The feature maps learned from these layers are called the joint feature maps, which come from each input pair to encode the relationship of two images.
Then they are sent into the full connected layers to calculate the similarity.
The joint feature maps hold the identifiable information of the input image pair that can represent the relationship of two images.
We use these joint feature maps to identify whether the input image pair is from the same person.
The classification loss in our network is the binary logistic regression loss, the same as the binary softmax loss in ~\cite{Li/cvpr2014deepreid,ahmed/cvpr2015improved}:

\begin{equation}
L_{cls}=-\sum_{i=1}^N{[(1-y)p(y=0|x)+yp(y=1|x)]}
\label{eq:clsloss}
\end{equation}

where $y\in\{0,1\}$. When the input pair is a positive pair (\eg ($A_1,A_2$)), $y=1$. On the contrary, $y=0$ for a negative pair (\eg $(A_1,B_2)$). $p(y|x)$ is the discrete probability distribution over two categories $y\in\{0,1\}$.

Our five convolutional layers are extended from the architecture of AlexNet \cite{krizhevsky/nips2012imagenet},
differing in that the size of each kernel in the third convolutional layer is $(512\times3\times3)$ instead of $(256\times3\times3)$ used in AlexNet.
In the train phase, the triplet loss optimises the first two convolutional layers
while the classification loss simultaneously trained all five convolutional layers including the first two.
In other words, the kernels of the first two layers are jointly optimised by two losses for extracting a global feature of each image.
The left three layers are mainly trained by the classification loss to obtain an identifiable feature for image pairs
to achieve the binary person identification.
In the test phase, only the classification task architecture (including the first two layers) is used.
The input two images are sent through five convolutional layers and three fully connected layers,
with the last layer predicting the similarity probability of a test pair.

\subsection{Cross-domain architecture}
\label{ssec:crossdomainarchitecture}

For most person ReID datasets, the size of data is too small to train a deep model.
The common way is to crop or mirror the images, which can increase the number of samples in datasets.
However, even with these augmentation processes, the total number of the samples is still far from the requirement of deep learning.
This problem is considered as a semi-supervised cross-domain issue in this paper.
In cross-domain transfer, the assumption is that two domains share the same task but the data distributions are different.
For example, in image classification, two domains would have the same category but the images contain different views or illuminations.
In our issue, the corresponding assumption is that two ReID datasets should share the same similarity function
while different variations caused by views or poses widely exist in images from two datasets.

In Fig.\ref{fig:framework}, the relationship of two images is reflected by the joint feature maps.
For two positive pairs from two different datasets, the learned similarity metrics for each of the pairs should ideally
lead to the same prediction results, \ie both of the pairs are matched pairs.
To achieve such a transfer, we propose to force the learned joint feature maps of positive pairs from two datasets closer than those of negative pairs.

The proposed cross-domain architecture is also shown in Fig.\ref{fig:framework},
which utilizes a contrastive loss~\cite{chopra/cvpr2005learning} to keep the two sets of joint feature maps of the same class as similar as possible during the training process.
The label for the two pairs is designed as following:

\begin{equation}
label_{p}=label_a\odot label_b
\label{eq:label}
\end{equation}

where $\odot$ means the XNOR operation, $label_a \in \{0,1\}$ is the label for a pair from source; $label_b \in \{0,1\}$ is the label for a pair from target; $label_{p}$ is the result after performing the XNOR operation between the labels of those two pairs. If the labels of the two pairs are the same (\ie $label_a$ and $label_b$ are the same), the contrastive loss will keep the two sets of the joint feature maps closer, and otherwise farther. The loss is as following:

\begin{equation}
\begin{aligned}
& L_{cts}=-\sum_{i=1}^N{[y\frac{1}{2}d_w^2+(1-y)\frac{1}{2}max(0,m-d_w)^2]} \\
& \phantom{XXXXXXX}d_w=\|F_a-F_b\|_2
\label{eq:contrastloss}
\end{aligned}
\end{equation}

where $y$ is the label of two pairs after the XNOR operation,
$F_a$ and $F_b$ are responses of the feature maps after the second fully connected layer from two datasets.

The training phase of the cross-domain architecture is also a multi-task process.
The softmax loss and the triplet loss are to do the re-identification task,
while the contrastive loss is employed to keep two sets of joint feature maps from the same class in two datasets as similar as possible.
After training, only the model on the target dataset will be reserved for testing.
The whole process can be considered as another kind of fine-tune operation using a cross-domain architecture.
The purpose is to use the joint feature maps learned on the auxiliary source dataset
to fine tune those on smaller target sets during training and boost the ReID performances.

It is worth noting that we don't force the feature maps of two completely different people, each from one of two
datasets, to be similar. Instead we ensure that the way in which image pairs are compared (encoded by the learned
weights on the joint feature maps) is similar and could be shared across the two datasets. That is the motivation
of importing the cross-domain architecture.

\section{Experiments}
\label{sec:experiments}

We conducts two sets of experiments: 1) to evaluate the proposed multi-task deep net (including single-task nets) and the cross-domain architecture; 2) to compare the proposed approach with state of the arts.

\subsection{Setup}
\subsubsection{Implementation and protocol.}
Our method is implemented using the Caffe framework \cite{jia2014caffe}.
All images are resized to $224\times224$ before being fed to network.
The learning rate is set to $10^{-3}$ consistently across all experiments.
For all the datasets, we horizontally mirror each image
and increase the dataset sizes fourfold.
We use a pre-trained AlexNet model (trained on Imagenet dataset~\cite{krizhevsky/nips2012imagenet})
to initialize the kernel weights of the first two convolutional layers.
Cumulative Matching Characteristics (CMC) curves are employed to measure the ReID performance. We report the single-shot results on all the datasets.

\subsubsection{Dataset and settings.} The experiment is conducted on one large dataset and four small datasets. The large dataset is CUHK03 \cite{Li/cvpr2014deepreid}, containing 13164 images from 1360 persons. We randomly select 1160 persons for training, 100 persons for validation and 100 persons for testing,
following exactly the same setting as \cite{Li/cvpr2014deepreid} and \cite{ahmed/cvpr2015improved}.
The four small datasets are CUHK01~\cite{li/accv2012human}, VIPeR~\cite{gray2007evaluating},
iLIDS~\cite{zheng/bmvc2009associating} and PRID2011~\cite{hirzer2011person}.
In CUHK01 dataset, we randomly choose only 100 persons for testing, and all the rest 871 persons are used for training.
For three other datasets, we randomly divide the individuals into two equal parts, with one used for training and the other for testing.
Specifically, in the PRID2011 dataset, besides 100 test individuals, there are another 549 people in the gallery.


\begin{table*}
\renewcommand{\arraystretch}{1.4}
\caption{The CMC performance of the state-of-the-art methods and different architectures in our method on five representative datasets. The bold indicates the best performance.}
\begin{center}
\resizebox{\linewidth}{!}{
\begin{tabular} {c|c||c|c|c||c|c|c||c|c|c||c|c|c||c|c|c}
  \hline
  \hline
  \multirow{2}{*}{Method} & \multirow{2}{*}{Type} & \multicolumn{3}{|c||}{CUHK03} & \multicolumn{3}{|c||}{CUHK01} & \multicolumn{3}{|c||}{VIPeR} & \multicolumn{3}{|c||}{iLIDS} & \multicolumn{3}{|c}{PRID2011}\\
  \cline{3-17}
   & & $r=1$ & $r=5$ & $r=10$ & $r=1$ & $r=5$ & $r=10$ & $r=1$ & $r=5$ & $r=10$ & $r=1$ & $r=5$ & $r=10$ & $r=1$ & $r=5$ & $r=10$ \\
  \hline
  PRDC~\cite{zheng/cvpr2011person} & - & - & - & -& - & - & -& 15.70 & 38.40 & 53.90 & 37.80 & 63.70 & 75.10 & 4.50 & 12.60 & 19.70\\
  SDALF~\cite{farenzena/cvpr2010person} & - & 5.60 & 23.45 & 36.09 & 9.90 & 41.21 & 56.00 & 19.87 & 38.89 & 49.37 & - & - & - & - & - & - \\
  ITML~\cite{davis/icml2007information} & - & 5.53 & 18.89 & 29.96 & 17.10 & 42.31 & 55.07 & - & - & - & 29.00 & 54.00 & 70.50 & 12.00 & - & 36.00 \\
  eSDC~\cite{zhao/cvpr2013unsupervised} & - & 8.76 & 24.07 & 38.28 & 22.84 & 43.89 & 57.67 & 26.31 & 46.61 & 58.86 & - & - & - & - & - & - \\
  KISSME~\cite{Koestinger/cvpr2012scale} & - & 14.17 & 48.54 & 52.57 & 29.40 & 57.67 & 62.43 & 19.60 & 48.00 & 62.20 & 28.50 & 55.30 & 68.70 & 15.00 & - & 39.00 \\
  FPNN~\cite{Li/cvpr2014deepreid} & Cls & 20.65 & 51.00 & 67.00 & 27.87 & 64.00 & 77.00 & - & - & - & - & - & - & - & - & -\\
  mFilter~\cite{Zhao/cvpr2014midlevel} & - & - & - & - & 34.30 & 55.00 & 65.30 & 29.11 & 52.34 & 65.95 & - & - & - & - & - & - \\
  kLFDA~\cite{xiong/eccv2014person} & - & 48.20 & 59.34 & 66.38 & 42.76 & 69.01 & 79.63 & 32.33 & 65.78 & 79.72 & 39.80 & 65.30 & 77.10 & 22.40 & 46.60 & 58.10 \\
  DML~\cite{yi/icpr2014deep} & Cls & - & - & - & - & - & - & 34.40 & 62.15 & 75.89 & - & - & -  & 17.90 & 37.50 & 45.90 \\
  IDLA~\cite{ahmed/cvpr2015improved} & Cls & 54.74 & 86.50 & 94.00 & 65.00 & 89.50 & 93.00 & 34.81 & 63.32 & 74.79 & - & - & - & - & - & - \\
  SIRCIR~\cite{sircir2016cvpr} & Cls/Rnk & 52.17 & 85.00 & 92.00 & 72.50 & 91.00 & 95.50  & 35.76 & 67.00 & 82.50 & - & - & - & - & - & - \\
  DeepRanking~\cite{chen2015deep} & Rnk & - & - & - & 70.94 & 92.30 & 96.90 & 38.37 & 69.22 & 81.33 & - & - & - & - & - & - \\
  DeepRDC~\cite{ding2015deep} & Rnk & - & - & - & - & - & - & 40.50 & 60.80 & 70.40 & 52.10 & 68.20 & 78.00 & - & - & - \\
  NullReid~\cite{null2016cvpr} & - & 58.90 & 85.60 & 92.45 & 64.98 & 84.96 & 89.92 & 42.28 & 71.46 & 82.94 & - & - & - & 29.80 & 52.90 & \textbf{66.00} \\
  SiameseLSTM~\cite{lstm2016cvpr} & Cls & 57.30 & 80.10 & 88.30 & - & - & - & 42.40 & 68.70 & 79.40 & - & - & - & - & - & - \\
  Ensembles~\cite{paisitkriangkrai/cvpr2015learning} & - & 62.10 & 89.10 & 94.30 & 53.40 & 76.30 & 84.40 & 45.90 & \textbf{77.50} & \textbf{88.90} & 50.34 & 72.00 & 82.50 & 17.90 & 40.00 & 50.00 \\
  GatedSiamese~\cite{gated2016cvpr} & Cls & 68.10 & 88.10 & 94.60 & - & - & - & 37.80 & 66.90 & 77.40 & - & - & - & - & - & -  \\
  ImpTrpLoss~\cite{imptrp2016cvpr} & Rnk & - & - & - & 53.70 & 84.30 & 91.00 & \textbf{47.80} & 74.70 & 84.80 & \textbf{60.40} & \textbf{82.70} & \textbf{90.70} & 22.00 & - & 47.00 \\
  \hline
  \hline
  MTDnet-rnk & Rnk & 60.13 & 90.51 & 95.15 & 63.50 & 80.00 & 89.50 & 28.16 & 52.22 & 65.19 & 41.04 & 69.94 & 78.61 & 22.00 & 41.00 & 48.00 \\
  MTDnet-cls & Cls & 68.35 & 93.46 & 97.47 & 76.50 & 94.00 & 97.00 & 44.30 & 69.94 & 81.96 & 54.34 & 73.41 & 86.13 & 28.00 & 50.00 & 60.00 \\
  MTDnet-trp & Cls+Rnk & 66.03 & 84.81 & 89.87 & 66.00 & 84.00 & 91.50 & 34.81 & 60.13 & 72.78 & 46.82 & 72.83 & 81.50 & 26.00 & 49.00 & 57.00 \\
  MTDnet & Cls+Rnk & \textbf{74.68} & \textbf{95.99} & \textbf{97.47} & 77.50 & 95.00 & 97.50 & 45.89 & 71.84 & 83.23 & 57.80 & 78.61 & 87.28 & \textbf{32.00} & 51.00 & 62.00 \\
  MTDnet-aug & Cls+Rnk & - & - & - & 75.50 & 93.50 & 97.00 & 43.35 & 70.25 & 78.48 & 54.91 & 74.57 & 84.97 & 27.00 & 46.00 & 59.00 \\
  MTDnet-cross & Cls+Rnk & - & - & - & \textbf{78.50} & \textbf{96.50} & \textbf{97.50} & 47.47 & 73.10 & 82.59 & 58.38 & 80.35 & 87.28 & 31.00 & \textbf{54.00} & 61.00 \\
  \hline
\end{tabular}}
\end{center}
\label{table:exp1}
\end{table*}

\subsection{Results for the multi-task network}
\label{ssec:experiment1}

\subsubsection{Multi vs. single task.} Results of CMCs with different rank accuracies are shown in Table.~\ref{table:exp1}.
The proposed multi-task network (Fig.~\ref{fig:framework}) is denoted by \textit{MTDnet}.
As \textit{MTDnet} adopts the classification loss for testing, we give results using the ranking loss for testing with the same model (denoted by \textit{MTDtrp}). It's obvious that the performance of \textit{MTDnet} is much better than \textit{MTDtrp} which implies the last three convolutional layers trained with the classification loss indeed provide a great help to increase the person ReID performance.
The results of the single-task networks using the triplet ranking loss (denoted by \textit{MTDnet-rnk}) and the binary classification loss (denoted by \textit{MTDnet-cls}) individually are also provided.
It is worth noting that, for a fair comparison, the architecture of \textit{MTDnet-rnk} network is expanded into containing five convolutional layers plus three fully connected layers as AlexNet~\cite{krizhevsky/nips2012imagenet} instead of the two convolutional layers shown in Fig.~\ref{fig:framework}, \ie the number of layers in two single-task networks is the same. The similarity of two images in \textit{MTDnet-rnk} is computed with the Euclidean distance. On CUHK03, our multi-task network (\textit{MTDnet}) achieves a rank-1 accuracy of 74.68\% and is much better than either \textit{MTDnet-cls} or \textit{MTDnet-rnk}, which indicates the complementarity of two tasks and the effectiveness of jointly optimizing. On four small datasets, our multi-task network consistently outperforms each of two single-task nets (\textit{MTDnet-cls} and \textit{MTDnet-rnk}).

\subsubsection{Cross-domain architecture.} We compare the cross-domain architecture (\textit{MTDnet-cross})
with the original multi-task network (\textit{MTDnet}) on four small datasets.
In this experiment, CUHK03 is considered as the dataset from the source domain,
while each of the four small dataset is from the target domain.
Therefore, the knowledge transfer is from CUHK03 to each of the four small datasets.
The results of \textit{MTDnet} on four small datasets is obtained by fine tuning the CUHK03 trained model on each small dataset.
In the cross-domain architecture, both the target domain network and the source domain network are initialized using the model trained on CUHK03.
And in test phase, only the target domain network is used to compute results.
Relevant preformance are shown in Table.\ref{table:exp1}.
It's obvious that almost all results of the cross-domain architecture are better than those of \textit{MTDnet}, which demonstrates the effectiveness of the cross-domain architecture.
We also import another network (\textit{MTDnet-aug}) which simply adds the source data into the target dataset directly and combined them as an augmented dataset for the target dataset training.
It's clear that the results of our cross-domain architecture are better than those of \textit{MTDnet-aug}.
The models trained with the augmented data (\textit{MTDnet-aug}) are even worse compared with \textit{MTDnet},
which suggests that the direct combination of the source and target datasets is not helpful but disruptive for the training in the target dataset.

\subsection{Comparison with the state of the arts}
\label{ssec:experiment2}
We compare ours with representative ReID methods including 18 algorithms,
whichever have the results reported on at least one of the five datasets.
All of the results can be seen from Table. \ref{table:exp1}.
We have marked all the deep learning methods in the \textit{Type} column.
All the non-deep learning approaches are listed as ``-''.
\textit{Cls} indicates deep methods based on the classification loss,
while \textit{Rnk} are on the ranking loss.
SIRCIR method~\cite{sircir2016cvpr} offers the results on both the classification loss and the ranking loss.
But in its network, the losses are trained separately.
Its combination of two losses are only on the score level,
while we jointly optimize two losses in one network and train them simultaneously.
Most of these deep methods are in the top performance group among all of the methods considered.
It is noted that our results are better than most approaches above,
which further confirms that jointly optimizing the two losses has a clear advantage over a single loss.
Under the rank-1 accuracy, our multi-task network outperforms all existing person ReID algorithms on CUHK03, CUHK01 and PRID2011.
ImpTrpLoss~\cite{imptrp2016cvpr} provides the best rank-1 performance on VIPeR and iLIDS.
We can see our results are comparable with its, and much better on other datasets.

\section{Conclusion}
\label{sec:conclusion}

In this paper, a multi-task network has been proposed for person ReID,
which integrates the classification and ranking tasks together in one network and takes the advantage of their complementarity.
In the case of having small target datasets, a cross-domain architecture has been further introduced
to fine tune the joint feature maps and improve the performance.
The results of the proposed network have outperformed almost all state-of-the-art methods compared on both large and small datasets.

\section{Acknowledgement}
This work is funded by the National Natural Science Foundation of China (Grant No. 61322209, Grant No. 61673375 and Grant No. 61403383), and the International Partnership Program of Chinese Academy of Science, Grant No. 173211KYSB20160008.

{\small
\bibliographystyle{aaai}
\bibliography{refs}
}

\end{document}